%% file: main.tex
\documentclass[conference]{IEEEtran}
\usepackage{graphicx}
\usepackage{times}
\usepackage{subcaption}
\usepackage{float}
\usepackage{booktabs}      % for \toprule, \midrule, \bottomrule
\usepackage{multirow}      % for \multirow
\usepackage[table]{xcolor} % for \rowcolor in tables
\usepackage{amsmath}
\usepackage{amsfonts}
\usepackage{amssymb}
\usepackage[numbers]{natbib}
\usepackage{multicol}
\usepackage[bookmarks=true]{hyperref}

\begin{document}

\title{Recurrent-Depth VLA: Implicit Test-Time Compute Scaling of Vision–Language–Action Models via Latent Iterative Reasoning
}

% You will get a Paper-ID when submitting a pdf file to the conference system
\author{
    \IEEEauthorblockN{
        Yalcin Tur$^{1}$,
        Jalal Naghiyev$^{2,\dagger}$,
        Haoquan Fang$^{3,4,\dagger}$,
        Wei-Chuan Tsai$^{3}$ \\
        Jiafei Duan$^{3,4,*}$,
        Dieter Fox$^{3,4,*}$,
        Ranjay Krishna$^{3,4,*}$
    }
    \IEEEauthorblockA{$^{1}$Stanford University \quad $^{2}$Technical University of Munich}
    \IEEEauthorblockA{$^{3}$University of Washington \quad $^{4}$Allen Institute for Artificial Intelligence}
    \IEEEauthorblockA{$^{\dagger}$Co-second authors}
    \IEEEauthorblockA{$^{*}$Co-advising}
    \IEEEauthorblockA{\href{http://rd-vla.github.io/}{\textcolor{blue}{rd-vla.github.io}}}
}

%\author{\authorblockN{Michael Shell}
%\authorblockA{School of Electrical and\\Computer Engineering\\
%Georgia Institute of Technology\\
%Atlanta, Georgia 30332--0250\\
%Email: mshell@ece.gatech.edu}
%\and
%\authorblockA{Twentieth Century Fox\\
%\and
%\authorblockN{James Kirk\\ and Montgomery Scott}
%\authorblockA{Starfleet Academy\\
%San Francisco, California 96678-2391\\

% avoiding spaces at the end of the author lines is not a problem with
% conference papers because we don't use \thanks or \IEEEmembership

%Homer Simpson\authorrefmark{2},
%Montgomery Scott\authorrefmark{3} and
%Eldon Tyrell\authorrefmark{4}}
%Telephone: (800) 555--1212, Fax: (888) 555--1212}
%\authorblockA{\authorrefmark{4}Tyrell Inc., 123 Replicant Street, Los Angeles, California 90210--4321}}

\maketitle

\begin{abstract}
Current Vision–Language–Action (VLA) models rely on fixed computational depth, expending the same amount of compute on simple adjustments and complex multi-step manipulation. While Chain-of-Thought (CoT) prompting enables variable computation, it scales memory linearly and is ill-suited for continuous action spaces. We introduce Recurrent-Depth VLA (RD-VLA), an architecture that achieves computational adaptivity via latent iterative refinement rather than explicit token generation. RD-VLA employs a recurrent, weight-tied action head that supports arbitrary inference depth with a constant memory footprint. The model is trained using truncated backpropagation through time (TBPTT) to efficiently supervise the refinement process. At inference, RD-VLA dynamically allocates compute using an adaptive stopping criterion based on latent convergence. Experiments on challenging manipulation tasks show that recurrent depth is critical: tasks that fail entirely (0\% success) with single-iteration inference exceed 90\% success with four iterations, while simpler tasks saturate rapidly. RD-VLA provides a scalable path to test-time compute in robotics, replacing token-based reasoning with latent reasoning to achieve constant memory usage and up to 80× inference speedup over prior reasoning-based VLA models.
 \end{abstract}

\IEEEpeerreviewmaketitle
\input{section/intro}
\input{section/relatedwork}
\input{section/methods}
\input{section/experiments}

\input{section/conclusion}

\bibliographystyle{plainnat}
\bibliography{references}
\include{section/appendix}

\end{document}

%% file: section/intro.tex
\section{Introduction}
Humans do not operate with a fixed computational budget. When performing trivial maneuvers, such as adjusting a grip or nudging an object, we rely on near-reflexive, low-effort responses. However, when environments become cluttered or actions require long-horizon foresight, human cognition adaptively reallocates resources—slowing down to process sensory evidence and refine internal models before acting \cite{verguts2015adaptive}. This ability to scale "compute" to task complexity is a hallmark of efficient reasoning that remains a significant challenge for modern robotics. Currently, most Vision-Language-Action (VLA) models \cite{kim2024openvla,black2024pi_0,lee2025molmoact} are limited to a fixed computational depth, processing every control step with the same number of parameters regardless of difficulty; a simple gripper adjustment receives the same compute as high-precision navigation in a cluttered space.

While generalist VLAs built upon Multimodal Large Language Models (MLLMs) exhibit impressive visual understanding, they often remain brittle and fail to fully leverage the latent reasoning capabilities of their backbone models. To bridge this gap, reasoning-centric VLAs \cite{huang2025thinkact,lee2025molmoact,zhao2025cot,zawalski2024robotic} have emerged. These models incorporate intermediate thinking processes—typically via supervised Chain-of-Thought (CoT) which categorized into textual reasoning (generating pseudo-CoT labels) or visual reasoning (generating sub-goal images or 2D traces) \cite{zheng2024tracevla,qu2025spatialvla,lee2025molmoact}. However, a fundamental limitation persists: these models perform reasoning at the token level. By tethering the "thinking" process to specific task settings and explicit sequences, these methods are often inefficient or misaligned with the requirements of continuous robotic control. But unlike language modeling, reasoning for physical manipulation is inherently subtler and less conducive to tokenization; consequently, attempting to verbalize these dynamics carries significant overhead, requiring the curation of custom reasoning datasets and scaling memory linearly with the length of the reasoning chain.

% One approach to achieving variable computational depth is Chain-of-Thought (CoT) prompting \cite{wei2022chain}, where models verbalize intermediate reasoning steps as tokens. Following its success in language modeling, prior work has adapted CoT for visuomotor policies by prompting models to generate auxiliary tokens such as sub-goal descriptions and categorical thought chains \cite{zawalski2024robotic}, spatial-temporal visual traces \cite{zheng2024tracevla}, or 2D/3D geometric reasoning primitives \cite{lee2025molmoact}. Unlike language modeling, reasoning for physical manipulation is inherently subtler and less conducive to tokenization; consequently, attempting to verbalize these dynamics carries significant overhead, requiring the curation of custom reasoning datasets and scaling memory linearly with the length of the reasoning chain.

Beyond these constraints, a fundamental architectural limitation persists: these models perform reasoning in the output space. This requires iterative transitions between the model’s high-dimensional latent space and a discretized, low-bandwidth output space (e.g., text, depth bins, or coordinates). This cyclical re-projection is fundamentally non-ideal, as it forces the model to collapse continuous internal states into lossy, discretized representations only to re-encode them for subsequent reasoning steps. The resulting information bottlenecks and quantization noise inherently limit reasoning fidelity, constraining the model's deliberation to the resolution of its output vocabulary rather than the continuous physical dynamics of the environment.

To address this, we look toward the iterative nature of biological systems. Human cognition is fundamentally characterized by recurrent neural dynamics, where the repeated recruitment of the same neural circuitry transforms initial sensory inputs into progressively refined, deliberative representations \cite{Lamme2000cb}. An architectural analogy in neural networks is the recursive reuse of layers, in which a model iteratively processes information until a refined decision is reached. This iterative process differs from Diffusion Policies \cite{chi2025diffusion}, which generate actions through multi-step refinement. Diffusion Policies operate by iteratively denoising an action trajectory in the output space. While effective for modeling multi-modal distributions, this process is fundamentally a generative sampling technique rather than a deliberative one: it refines the action signal but does not scale or enrich the underlying representation of the scene.

\begin{figure*}[h]
    \centering
    \includegraphics[width=1\linewidth]{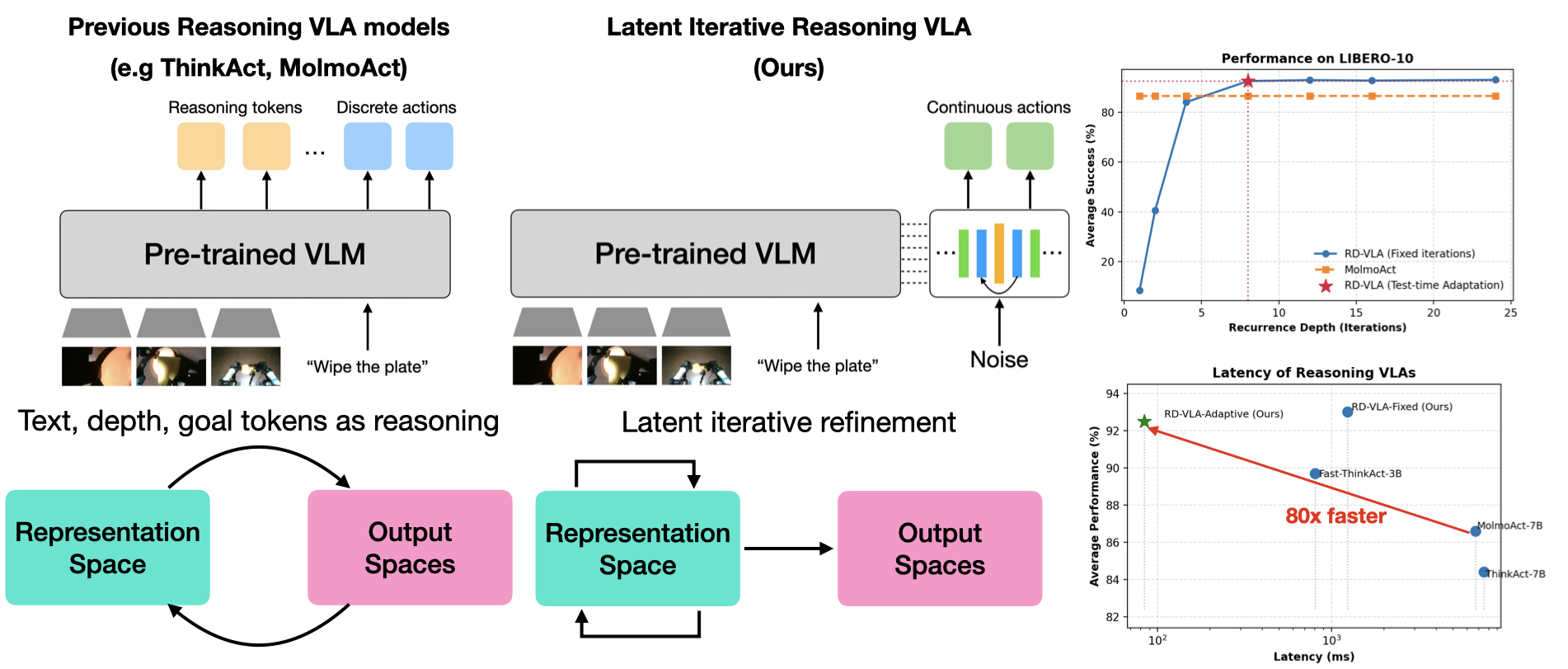}
    \caption{\textbf{Recurrent-Depth VLA.} (Left) Previous reasoning VLAs (e.g., ThinkAct, MolmoAct) generate explicit reasoning tokens in output space, requiring expensive autoregressive decoding. (Center) Our approach performs iterative refinement entirely in latent representation space, bypassing token generation overhead. (Right) RD-VLA achieves comparable performance to autoregressive reasoning baselines on LIBERO-10 while being substantially faster due to the efficiency of latent reasoning with adaptive compute.}

    \label{fig:1}
\end{figure*}

We introduce Recurrent-Depth VLA (RD-VLA), a new class of VLA model that enables adaptive test-time compute for visuomotor control. Unlike existing reasoning-centric methods, our approach enables adaptive behavior emerges naturally from its recurrent architecture, eliminating the need for curated reasoning traces or explicit supervision of iteration counts. Inspired by recent advances in latent reasoning and looped transformer architectures \cite{zhu2025scalinglatentreasoninglooped,geiping2025scalingtesttimecomputelatent}, we extend the principle of latent iterative refinement to the VLA domain. Our core insight is that robotic reasoning can occur entirely within the representation space, bypassing the need to decode intermediate tokens or perform computationally expensive denoising iterations in action space. RD-VLA achieves this by decomposing the action head into three stages: an initial encoding stack, a weight-tied recurrent block for iterative refinement, and a final projection layer. By recursively updating hidden states through the shared recurrent block, the model progressively refines its internal representations within a fixed-dimensional latent space. This design enables arbitrarily deep computation at test time while maintaining a constant memory footprint, with minimal parameter overhead due to weight reuse.

When coupled with an adaptive stopping criterion (Figure~\ref{fig:adaptive_complexity}), RD-VLA dynamically allocates computation on a per-sample basis, yielding a scalable and resource-efficient framework for high-precision robotic control at inference. To our knowledge, RD-VLA is the first VLA model to support scaling test-time computation through implicit latent-space reasoning via a weight-tied recurrent core. We further observe task-dependent convergence behavior: the required unrolled depth emerges naturally from the execution context, reflecting the underlying complexity of the action being performed. This property enables adaptive inference-time stopping and execution schedules that modulate the action horizon based on the model’s internal reasoning trajectory, resulting in inference speeds up to an order of magnitude faster than prior action-reasoning approaches. Empirically, RD-VLA outperforms strong end-to-end VLAs and all reasoning-centric VLAs on the LIBERO benchmark \cite{liu2023liberobenchmarkingknowledgetransfer}, achieving 93.0\% success with fixed iterations and 92.5\% with uncertainty-based adaptive computation. On the CALVIN benchmark \cite{mees2022calvinbenchmarklanguageconditionedpolicy}, RD-VLA achieves an average task length of 3.39 and a task-5 success rate of 45.3\%, demonstrating strong long-horizon generalization. Finally, RD-VLA transfers effectively to real-world robotic settings, achieving robust performance on challenging long-horizon tasks such as bread toasting and towel folding.

\begin{figure*}[h]
    \centering
    \includegraphics[width=1\linewidth]{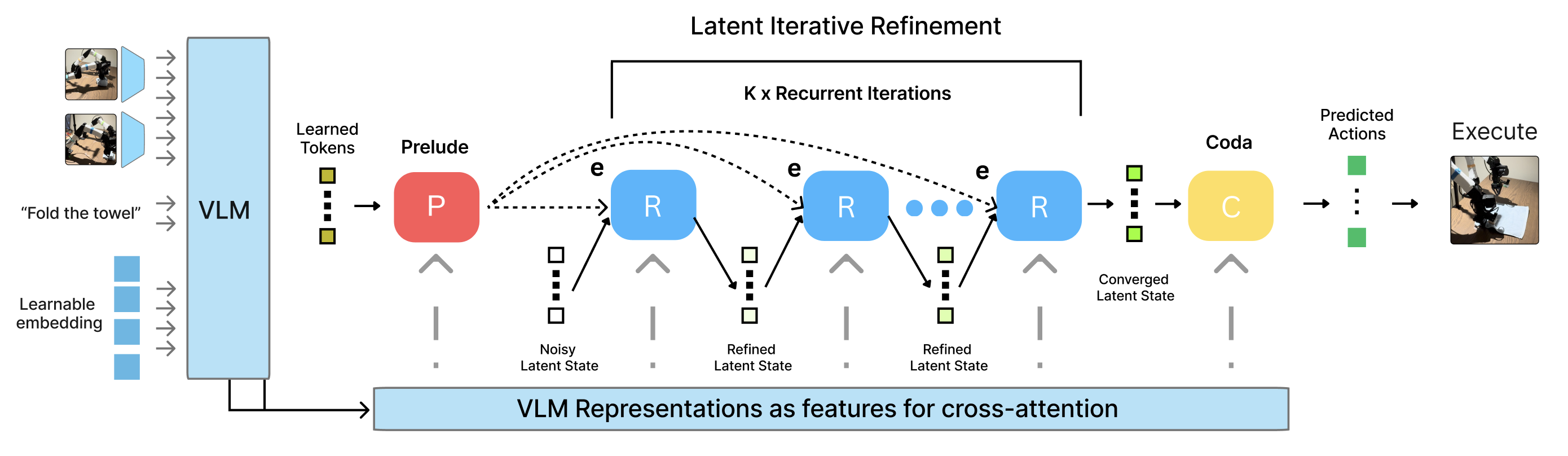}
    \caption{\textbf{Recurrent-Depth VLA Architecture.} The Prelude (P) grounds learned queries via cross-attention to mid-layer VLM features. The weight-tied Recurrent Core (R) iteratively refines a noisy latent scratchpad over $K$ iterations, cross-attending to final-layer VLM representations and proprioception. The Coda (C) decodes the converged state into actions. Recurrence depth $K$ adapts dynamically at inference based on task complexity.}

    \label{fig:rd-vla-architecture}
\end{figure*}
% \section{Method}
% \label{sec:methods}

% Across experiments on the LIBERO \cite{liu2023liberobenchmarkingknowledgetransfer} and CALVIN \cite{mees2022calvinbenchmarklanguageconditionedpolicy} benchmarks, as well as real-world robotic tasks, we demonstrate the effectiveness of latent compute scaling. We report:
% \begin{enumerate}
%     \item \textbf{LIBERO Benchmark:} Achieved a score of \textbf{93.0} with fixed iterations, and \textbf{92.1} with uncertainty-based adaptive computation.
%     \item \textbf{CALVIN Benchmark:} Recorded an average length of \textbf{3.39} on the CALVIN ABC$\to$D split.
%     \item \textbf{Real-World Deployment:} Demonstrated strong performance on long-horizon tasks, including toasting bread and folding towels.
% \end{enumerate}
% Although simple tasks often converge within a single iteration, more complex manipulations that fail under shallow recurrence achieve near-perfect success as test-time computation increases without increasing memory usage.

%We need to connect terminology and unify it% TODO CONNECT TODO ADD REFERENCES TO ADAPTIVE

%% file: section/relatedwork.tex
\section{Related Work}

\subsection{Vision-Language-Action Models}
Large language models (LLMs) \cite{achiam2023gpt, dubey2024llama, jiang2024mixtral} and vision language models (VLMs) \cite{liu2023llava, wang2024qwen2, karamcheti2024prismatic, zhu2023minigpt} have shown strong problem-solving capabilities and deep semantic understanding of visual and textual data. Leveraging large-scale Internet data, they generalize to unseen tasks, though acquiring comparable real-world robotic data remains challenging. Despite advances in assembling large robotic datasets \cite{ebert2021bridge, mendonca2023structured}, the Open X-Embodiment dataset \cite{o2024open} is the largest, featuring 22 robots, 527 skills, and 160,266 tasks from 21 institutions.

Using this dataset, Transformer-based general robot policies like Octo \cite{team2024octo} and RT-1 \cite{brohan2022rt} were trained. However, models trained from scratch for robotic tasks struggled to generalize to new environments, tasks, and objects. RT-2 \cite{zitkovich2023rt} addressed these issues by using a large (55B-parameter) pretrained vision-language model to generate actions, while OpenVLA \cite{kim2024openvla} emerged as a leading open-source alternative. Building on these, $\pi_{0}$ \cite{black2024pi_0} introduced a generalist policy using flow-matching for multi-modal actions, with $\pi_{0.5}$\cite{intelligence2025pi_0_5} providing a more efficient, high-performance iteration for real-time deployment.

\subsection{Reasoning and Efficient-Compute VLA Models}
As Vision-Language-Action (VLA) models scale in parameter count, balancing the tradeoff between computational demand and the real-time requirements of robotic control has become increasingly critical. Recent research has split into two primary directions: optimizing the efficiency of existing backbones and introducing structured reasoning stages to move beyond reactive policies.

To mitigate the high latency of large Transformer backbones, several works explore dynamic resource allocation. TinyVLA \cite{wen2025tinyvla} focuses on data-efficient distillation to create high-performance, small-scale models. To reduce redundant computation in invariant visual scenes, VLA-Cache \cite{xu2025vla} introduces an adaptive token caching mechanism that reuses textual and visual features across control steps. Alternatively, DeeR-VLA \cite{yue2025deer} treats model depth as a dynamic variable, activating model segments based on task difficulty. This enables rapid early-exiting during trivial movements while preserving full capacity for high-entropy manipulations.

A second direction aims to improve performance by grounding actions in explicit reasoning via mid-level representations. Mobility-VLA \cite{chiang2024mobility} employs long-context VLMs to reason over topological graphs for navigation, while TraceVLA \cite{zheng2024tracevla} utilizes visual trace prompting to enhance spatial-temporal awareness. Several recent works have integrated explicit reasoning directly into the control loop. Embodied Chain of Thought (ECoT) \cite{zawalski2024robotic} leverages embodied chain-of-thought to generate textual justifications before action emission. ThinkAct \cite{huang2025thinkact} utilizes reinforced visual latent planning, and MolmoAct \cite{lee2025molmoact} introduces Action Reasoning Models (ARMs) that generate depth-aware perception tokens and editable trajectories. Similarly, CoT-VLA \cite{zhao2025cot} demonstrates that visual chain-of-thought reasoning about spatial constraints and object relationships significantly improves generalization in out-of-distribution environments.

\subsection{Recurrent Transformers}
\begin{figure}[ht]
    \centering
    \includegraphics[width=1.0\linewidth]{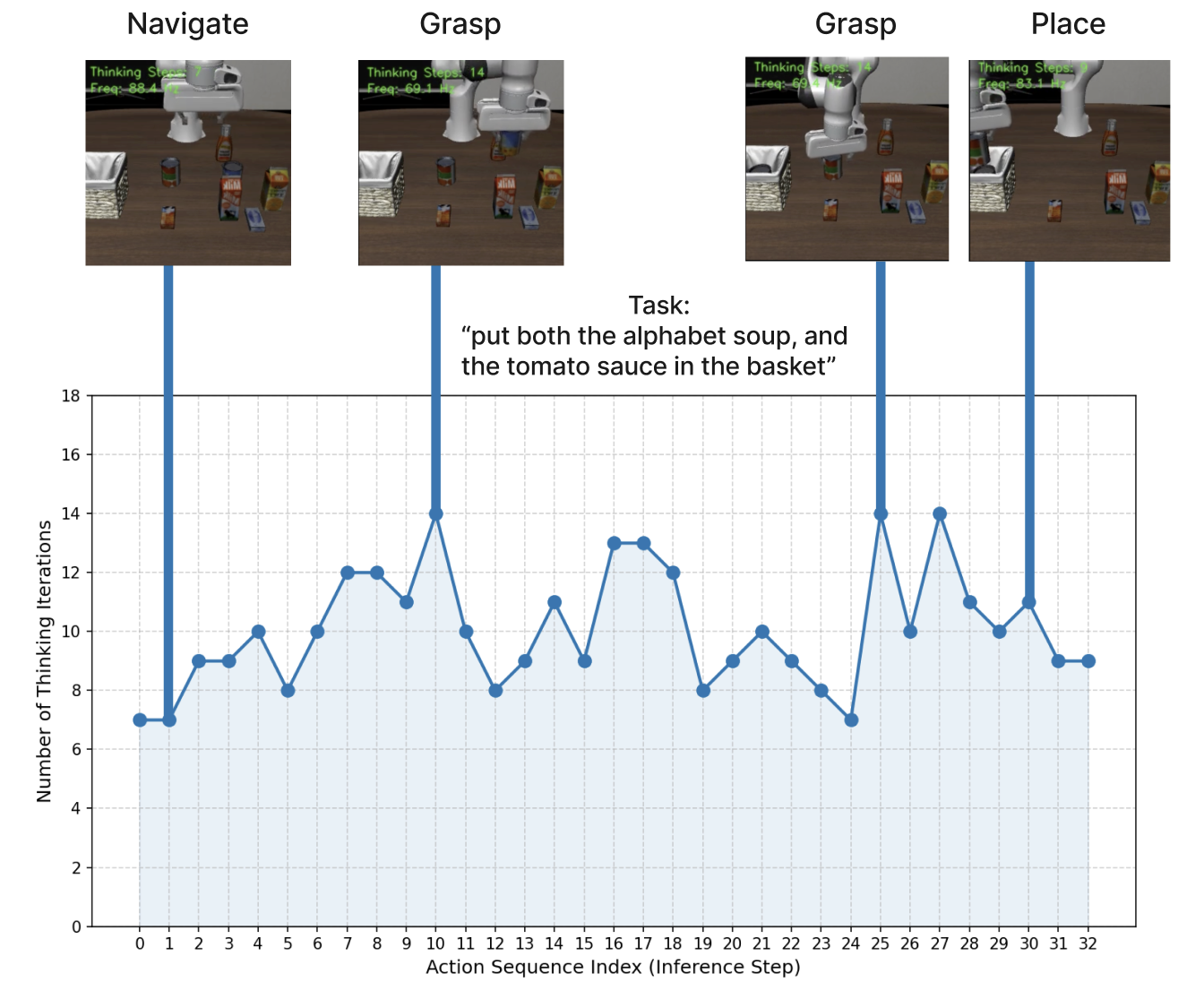}
    \caption{\textbf{Case study for adaptive computation.} In a LIBERO rollout, the model dynamically selects different numbers of iterations before terminating, depending on the execution state. It uses fewer iterations (7–9) at steps 1 and 30, which correspond to simpler motions like navigation and placing, and more iterations (about 14) at steps 10 and 25, where the actions are more complex, such as grasping.}
    \label{fig:adaptive_complexity}
\end{figure}

Recurrent Transformers is an architecture where all or some portion of layers in transformers are recurred, which means that representation of the later layer gets reinjected back into the earlier layers. The initial work focused on recurring only one transformer layer \cite{dehghani2019universaltransformers}, while later works explored various architectures and methods to train recurrent transformers \cite{geiping2025scalingtesttimecomputelatent, gatmiry2024roledepthloopingincontext,hao2025traininglargelanguagemodels,mcleish2025teachingpretrainedlanguagemodels}. Although various research mostly focused on inspecting inductive biases of recurrent architectures on toy scales, and in algorithmic setting \cite{wang2025hierarchicalreasoningmodel, jolicoeurmartineau2025morerecursivereasoningtiny, saunshi2024inductivebiasstackingimproving, mcleish2024transformersarithmeticrightembeddings}, recent results had shown that these model could scale to Foundation Models sizes \cite{geiping2025scalingtesttimecomputelatent,mcleish2025teachingpretrainedlanguagemodels,zhu2025scalinglatentreasoninglooped}. Recurrent transformers allow models to scale computation by repeating layers, enabling several useful properties: test-time scaling, where the number of computation steps can be increased at inference; uncertainty quantification for test-time scaling, since operating at the representation level makes it possible to define metrics that reflect model confidence; adaptive compute \cite{geiping2025scalingtesttimecomputelatent}, where prior work uses measures such as cosine distance and KL divergence between representations across recurrent iterations to adjust the number of computation steps based on uncertainty, and we illustrate this behavior in simulation in Figure \ref{fig:adaptive_complexity}; and no need for specialized CoT data, since recurrent transformers provide an architectural mechanism for iterative reasoning without requiring curated chain-of-thought supervision, which can be particularly difficult to obtain in robotics.

%Don't see major issues % TODO EDIT 4 POINTS IN THE END

%% file: section/methods.tex
% \begin{figure*}
%     \centering
%     \includegraphics[width=1\linewidth]{figure2.png}
%     \caption{\textbf{Recurrent-Depth VLA Architecture.} The Prelude (P) grounds learned queries via cross-attention to mid-layer VLM features. The weight-tied Recurrent Core (R) iteratively refines a noisy latent scratchpad over $N$ iterations, cross-attending to final-layer VLM representations and proprioception. The Coda (C) decodes the converged state into actions. Recurrence depth $N$ adapts dynamically at inference based on task complexity.}

%     \label{fig:rd-vla-architecture}
% \end{figure*}
% \section{Method}
% \label{sec:methods}
\section{Method}
\label{sec:methods}
We introduce \textbf{Recurrent-Depth VLA (RD-VLA)}, a framework designed to decouple computational depth from the fixed architectural constraints of pretrained vision-language backbones. While standard VLA architectures typically utilize fixed-depth MLP heads or compute-intensive iterative processes in the output space such as diffusion or flow-matching action heads \cite{black2024pi_0}, RD-VLA shifts the computational burden to a weight-tied recurrent transformer core operating entirely within a continuous latent manifold. This design allows the model to scale its test-time compute by unrolling the recurrent block to an arbitrary depth $r$, enabling dynamic allocation of compute based on task complexity (Figure \ref{fig:adaptive_complexity}).

\subsection{Architectural Backbone and Token Flow}
\label{sec:methods:Backbone}

The RD-VLA action head is a modular framework designed to be backbone-agnostic, capable of being integrated with any Vision-Language Model (VLM) that produces dense latent representations. For the purposes of this work, we instantiate the framework using a VLM based on Qwen2.5-0.5B \cite{qwen2, qwen2.5} from MiniVLA \cite{belkhale2024minivla}, which utilizes the Prismatic \cite{karamcheti2024prismatic} training recipe with a frozen DINOv2 \cite{oquab2023dinov2} and SigLIP \cite{zhai2023sigmoid} fused vision encoder from MiniVLA \cite{belkhale2024minivla}.

The frozen vision encoder generates $256$ vision tokens per image observation ($512$ for wrist and the main camera), which are projected into a Qwen2-0.5B (24-layer) LLM backbone fine-tuned via LoRA. Following the architectural design principles of OpenVLA-OFT \cite{kim2025fine}, we augment the VLM input sequence with a set of $64$ dedicated learned latent embeddings. These tokens serve as specialized grounded placeholders that attend to the multi-modal context during the LLM's forward pass. 

After the VLM execution, we extract the hidden states and partition them into two distinct sets:
\begin{itemize}
    \item \textbf{Task/Vision representations} ($h_{vis} \in \mathbb{R}^{512 \times D}$): Capturing spatial and semantic scene information.
    \item \textbf{Latent-specific representations} ($h_{lat} \in \mathbb{R}^{64 \times D}$): Extracted from the latent token positions to provide compressed, task-aligned features.
\end{itemize}
These representations are combined to form a static conditioning manifold. Specifically, the recurrent head $R_\theta$ at each iteration $k$ attends via cross attention to a concatenated context vector $[h_{vis+lat}^{(24)}; p]$, effectively grounding the latent reasoning process in both observation and the high-level semantic tokens.
\subsection{Recurrent-Depth Architecture}
\label{sec:methods:Recurrent-Depth}

We introduce a framework that decouples computational depth from the fixed architectural constraints of pretrained vision-language backbones. While standard VLA architectures typically utilize fixed-depth heads, RD-VLA shifts the computational burden to a weight-tied recurrent transformer core operating within a continuous latent manifold. Following the Huggin approach \cite{geiping2025scalingtesttimecomputelatent}, we partition the architecture into a functional triplet: the \textbf{Prelude}, the \textbf{Recurrent Core}, and the \textbf{Coda}  (Fig.~\ref{fig:rd-vla-architecture}). The Prelude and Coda serve as non-recurrent interface layers that map representations into and out of a dedicated \textbf{latent manifold} optimized for iterative reasoning.

The process begins with the \textbf{Prelude} ($P_\phi$), a non-recurrent interface that consumes $K=8$ learned \textbf{queries}. First $K$ queries self attend to each other bidirectionally. Then by performing cross-attention over the VLM's \textbf{middle-layer} visual features $h_{vis+lat}^{(12)}$, the Prelude transforms these queries into a grounded latent foundation:
\begin{equation}
    S_{pre} = P_\phi(\text{Queries}, h_{vis+lat}^{(12)}) \in \mathbb{R}^{K \times D}
\end{equation}

Parallel to this, we initialize a \textbf{latent scratchpad} $S \in \mathbb{R}^{K \times D}$ from a high-entropy truncated normal distribution to serve as the evolving state for the reasoning process:
\begin{equation}
    S_0 \sim \text{TruncNormal}(0, \gamma_{init} \cdot \sigma_{init})
\end{equation}
This noisy initialization transforms $S$ into a blank workspace that the model must iteratively ``clean'' and refine. This ensures that the model learns a stable refinement operator rather than overfitting to a specific starting point.

\subsubsection{Latent Iterative Reasoning via Input Injection}
\label{sec:methods:Latent Iterative Reasoning}

The core iterative refinement occurs within the \textbf{Recurrent Core} ($R_\theta$), a weight-tied transformer block. To maintain representational stability and prevent the model from losing its grasp on the physical observations over long unrolls (representational collapse), we utilize a persistent \textbf{Input Injection} strategy. At every step $k$, the recurrent block observes both the current state of the scratchpad $S_{k-1}$ and the static foundation $S_{pre}$ provided by the Prelude.

Specifically, for each iteration $k=1 \dots r$, the previous scratchpad state $S_{k-1}$ is concatenated with the fixed $S_{pre}$ along the feature dimension to form a $2D$-dimensional context. This is mapped back to the manifold dimension via a learned adapter and normalized:
\begin{equation}
    x_k = \text{RMSNorm}\left( \gamma_{adapt} \cdot W_{adapt} [S_{k-1} ; S_{pre}] \right)
\end{equation}
where $W_{adapt} \in \mathbb{R}^{D \times 2D}$. The scratchpad state is then updated through the weight-tied recurrent block:
\begin{equation}
    S_k = R_\theta(x_k, [h_{vis+lat}^{(24)}; p])
\end{equation}

Similar to Prelude during this update, $R_\theta$ first performs bidirectional self-attention across K, and then does gated cross-attention where the queries are derived from $x_k$ and the keys/values are derived from a concatenated conditioning manifold $[h_{vis+lat}^{(24)}; p]$. This manifold consists of the 64 task-aligned latent tokens and 512 vision tokens from the VLM's final layer and the robot's current proprioception $p$.

\subsubsection{Coda and Action Projection}
\label{sec:methods:Coda}

Once the recurrence reaches the desired depth $r$, the converged scratchpad $S_r$ is processed by the non-recurrent \textbf{Coda} ($C_\psi$). The Coda performs the final decoding pass by moving the representation out of the latent manifold, attending to the self, and high-level VLM features ($h_{vis+lat}^{(24)}$). Finally, an \textbf{output projection} layer maps these refined features to the robot's action space:
\begin{equation}
    \mathbf{a} = W_{out} \cdot \text{RMSNorm}(C_\psi(S_r, [h_{vis}^{(24)}; h_{lat}^{(24)}; p]))
\end{equation}
where $W_{out}$ is the final linear layer producing the control commands $\mathbf{a}$. This architecture ensures that any intermediate state $S_k$ is a valid representation, while $S_{k+1}$ represents a strictly more refined iteration of the action plan.

\subsubsection{Training with Randomized Recurrence}
\label{sec:methods:Training}

To promote convergence to a steady state independent of initialization depth, we sample the number of iterations $N$ during training from a heavy-tailed log-normal Poisson distribution:
\begin{equation}
    \tau \sim \mathcal{N}(\ln(\mu_{rec}) - 0.125, \sigma^2), \quad N \sim \text{Poisson}(e^\tau) + 1
\end{equation}
where $\mu_{rec}=32$. We utilize \textbf{TBPTT}, where gradients are propagated through only the final $d=8$ iterations, while earlier steps are computed with gradients detached. This forces the network to learn to iteratively refine the scratchpad from any noisy initialization into a stable manifold, ensuring $S_{k+1}$ is a strictly better refinement of $S_k$. This enables the model to scale compute dynamically during inference without retraining.

\subsection{Adaptive Computation}
\label{subsec:adaptive_computation}

Leveraging the properties of the model, we implement an adaptive computation mechanism at inference. Rather than specifying a fixed iteration count, we utilize the model's own internal convergence as a proxy for reasoning certainty. 

We define a stopping criterion based on the Kullback-Leibler (KL) divergence between the action distributions of consecutive iterations. Approximating KL via Mean Squared Error (MSE) in the action space, the inference loop terminates at step $k^*$ when:
\begin{equation}
    ||\mathbf{a}_k - \mathbf{a}_{k-1}||^2_2 < \delta
\end{equation}
where $\mathbf{a}_k$ is the predicted action chunk at step $k$ and $\delta$ is a convergence threshold (e.g., $1e^{-3}$). This allows the model to self-regulate: terminating instantly for trivial movements while allocating extended compute for complex situations. 

\subsection{Adaptive Execution}
\label{subsec:adaptive_execution}

Adaptive computation determines how long to recur. At the same time adaptive execution determines how many actions to execute. We observe that instances requiring deep recurrence ($k^* > 8$) often correspond to states of high  uncertainty. In these regimes, executing a long horizon of actions is dangerous, as small errors in the initial plan compound over time.

We propose two strategies to couple the depth of reasoning with the execution of action.

\subsubsection{ Threshold-Based Adaptive Execution}
\label{subsec:Threshold-Based-adaptive}

This method modulates the execution horizon using a binary reasoning threshold $\tau$. We hypothesize that high iteration counts imply higher epistemic uncertainty. Consequently, if convergence requires $k^* > \tau$ steps, we truncate the horizon to a shorter duration $H_{short}$ to mitigate compounding errors, while retaining $H_{long}$ for confident predictions ($k^* \leq \tau$):
\begin{equation}
    H_{exec} = 
    \begin{cases} 
      H_{long} & \text{if } k^* \leq \tau \\
      H_{short} & \text{if } k^* > \tau
   \end{cases}
\end{equation}

\subsubsection{ Linear Decay Execution}
\label{subsec:Linear-decay}

To provide a continuous scaling mechanism, we implement a linear decay schedule that reduces the execution horizon inversely to the reasoning depth. Given a base iteration budget $\tau_{base}$, the horizon $H_{exec}$ decreases by one step for every additional iteration required to converge:
\begin{equation}
    H_{exec}(k^*) = \max\left(H_{min}, H_{max} - \max(0, k^* - \tau_{base})\right)
\end{equation}
This approach forces the agent to replan more frequently as computational demand increases, favoring safety over efficiency in complex states.

%% file: section/experiments.tex
\section{Experiments}
\label{sec:experiments}

\begin{table*}[!t]
    \centering
    \footnotesize
    \setlength{\tabcolsep}{3.4mm}
    \caption{Comparison on the LIBERO \cite{liu2023liberobenchmarkingknowledgetransfer} benchmark. \textbf{Bold\textsuperscript{*}} indicates the best performance, \textbf{Bold} the second best, and \underline{\textit{Italics}} the third best. ``Params'' denotes backbone scale in \textbf{B}illions. Here we compare three types of VLAs. \textbf{End-to-end (E2E) VLAs} directly predict robot actions without intermediate reasoning steps. \textbf{Token reasoning} VLAs first perform explicit reasoning by generating tokens before producing action outputs. Finally, \textbf{latent reasoning}, our approach, performs iterative reasoning in a latent space using a recurrent structure before emitting actions.}
    \label{ComparisonLIBERO}

    \begin{tabular}{l l c | c c c c | c}
        \toprule[1pt]
        \multicolumn{2}{l}{Method} & Params & Spat. & Obj. & Goal & Long & Avg. \\
        \midrule[0.5pt]

        \multirow{4}{*}{{\small \textbf{E2E VLAs}}}
        & SmolVLA \cite{shukor2025smolvla} & 2.2 & 93.0 & 94.0 & 91.0 & 77.0 & 88.8 \\
        & OpenVLA \cite{kim2024openvla} & 7 & 84.7 & 88.4 & 79.2 & 53.7 & 76.5 \\
        & WorldVLA \cite{cen2025worldvla} & 7 & 87.6 & 96.2 & 83.4 & 60.0 & 81.8 \\
        & $\pi_0$-FAST \cite{black2024pi_0} & 3 & 96.4 & 96.8 & 88.6 & 60.2 & 85.5 \\

        \midrule[0.5pt]

        \multirow{7}{*}{{\small \textbf{Token Reasoning}}}
        & CoT-VLA \cite{zhao2025cot} & 7 & 87.5 & 91.6 & 87.6 & 69.0 & 81.1 \\
        & FlowVLA \cite{zhong2025flowvla} & 8.5 & 93.2 & 95.0 & 91.6 & 72.6 & 88.1 \\
        & SpatialVLA \cite{qu2025spatialvla} & 4 & 88.2 & 89.9 & 78.6 & 55.5 & 78.1 \\
        & ThinkAct \cite{huang2025thinkact} & 7 & 88.3 & 91.4 & 87.1 & 70.9 & 84.4 \\
        & Fast-ThinkAct \cite{huang2026fast} & 3 & 92.0 & 97.2 & 90.2 & 79.4 & 89.7 \\
        & TraceVLA \cite{zheng2024tracevla} & 7 & 84.6 & 85.2 & 75.1 & 54.1 & 74.8 \\
        & MolmoAct \cite{lee2025molmoact} & 7 & 87.0 & 95.4 & 87.6 & 77.2 & 86.6 \\

        \midrule[0.5pt]

        \multirow{2}{*}{{\small \textbf{Latent Reasoning}}}
        & \textbf{RD-VLA (Fixed)}  & \textbf{0.5} & \textbf{92.0} & \textbf{99.0} & \textbf{96.0} & \textbf{84.8} & \textbf{93.0} \\
        & \textbf{RD-VLA (Adaptive)} & \textbf{0.5} & \textbf{88.6} & \textbf{98.8} & \textbf{96.8} & \textbf{85.8} & \textbf{92.5} \\

        \bottomrule[1pt]
    \end{tabular}
\end{table*}

We evaluate our approach in both simulation and real-world settings. Simulation experiments are conducted on two widely used manipulation benchmarks, LIBERO \cite{liu2023liberobenchmarkingknowledgetransfer} and CALVIN \cite{mees2022calvinbenchmarklanguageconditionedpolicy}, while real-world experiments are performed on a bimanual YAM manipulator. Our evaluation is designed to answer the following questions:
\begin{itemize}
    \item [\ref{subsec:fixed_recurrence}] How does our approach scale with recurrent computation?
    \item [\ref{subsec:task_dependent}] Is adaptive computation necessary for robotic manipulation tasks?
    \item [\ref{subsec:adaptive_experiments}] What is the most effective strategy for adaptive computation at inference time?
    \item [\ref{subsec:benchmark_comparison}] Is representation-level reasoning for action prediction more effective than token-level reasoning?
    \item [\ref{subsec:real_world}] How well does our approach perform in real-world deployment?
\end{itemize}

In this section, we evaluate Recurrent-Depth VLA (RD-VLA) across three dimensions: (1) the scaling behavior of latent recurrence on standard benchmarks, (2) the efficacy of adaptive test-time compute, and (3) performance comparisons against state-of-the-art baselines. We utilize the LIBERO \cite{liu2023liberobenchmarkingknowledgetransfer} and CALVIN \cite{mees2022calvinbenchmarklanguageconditionedpolicy} benchmarks for quantitative analysis and conduct real-world evaluations to assess physical robustness.

\subsection{Performance Scaling via Recurrent Computation}
\label{subsec:fixed_recurrence}

\begin{table}[!t]
    \centering
    \small
    \setlength{\tabcolsep}{1.8mm}
    \caption{Comprehensive Ablation of Recurrent-Depth VLA. We compare fixed computational depths against adaptive strategies (Binary, Linear Decay, and Pure KL). For adaptive runs, we report Mean Iterations ($\bar{k}$) and Standard Deviation ($\sigma$) to illustrate reasoning convergence.}
    \label{FullAblationTable}

    \resizebox{\linewidth}{!}{%
    \begin{tabular}{l l c c | c c c c | c}
        \toprule[1pt]
        \multicolumn{2}{l}{Strategy \& Threshold $\tau$} & $\bar{k}$ & $\sigma$ & Spatial & Object & Goal & Long & Avg. \\
        \midrule[0.5pt]

        \multirow{8}{*}{\textit{\shortstack{Fixed\\Recurrence}}} 
        & Rec=1  & 1.0 & 0.00 & 9.0  & 12.2 & 11.4 & 1.0  & 8.4  \\
        & Rec=2  & 2.0 & 0.00 & 38.0 & 61.2 & 47.6 & 15.0 & 40.5 \\
        & Rec=4  & 4.0 & 0.00 & 79.2 & 93.0 & 89.2 & 74.8 & 84.1 \\
        & Rec=8  & 8.0 & 0.00 & 93.0 & 97.8 & 94.2 & 85.2 & 92.6 \\
        & Rec=12 & 12.0 & 0.00 & 92.0 & 99.0 & 96.0 & 84.8 & 93.0 \\
        & Rec=16 & 16.0 & 0.00 & 91.6 & 99.0 & 95.2 & 85.2 & 92.8 \\
        & Rec=24 & 24.0 & 0.00 & 92.4 & 99.2 & 94.2 & 86.6 & \textbf{93.1} \\
        & Rec=32 & 32.0 & 0.00 & 91.4 & 98.8 & 93.8 & 84.4 & 92.1 \\
        
        \midrule[0.5pt]

        \multirow{6}{*}{\textit{\shortstack{Binary\\Adaptation}}} 
        & $\tau=1e^{-4}$ & 11.04 & 1.20 & 91.2 & 98.2 & 96.0 & 79.8 & 91.3 \\
        & $\tau=2e^{-4}$ & 9.71  & 1.11 & 90.6 & 97.2 & 96.0 & 80.8 & 91.2 \\
        & $\tau=5e^{-4}$ & 7.93  & 1.03 & 88.6 & 98.8 & 96.8 & 85.8 & \textbf{92.5} \\
        & $\tau=1e^{-3}$ & 6.61  & 0.89 & 88.6 & 97.8 & 94.6 & 84.8 & 91.5 \\
        & $\tau=5e^{-3}$ & 4.27  & 0.41 & 83.2 & 93.6 & 87.4 & 61.8 & 81.5 \\
        & $\tau=1e^{-2}$ & 3.36  & 0.19 & 74.6 & 88.6 & 81.6 & 43.6 & 72.1 \\

        \midrule[0.5pt]

        \multirow{6}{*}{\textit{\shortstack{Linear\\Decay}}} 
        & $\tau=1e^{-4}$ & 11.55 & 1.03 & 91.2 & 97.2 & 96.2 & 80.0 & 91.2 \\
        & $\tau=2e^{-4}$ & 9.86  & 1.10 & 91.2 & 98.4 & 96.6 & 82.0 & \textbf{92.1} \\
        & $\tau=5e^{-4}$ & 7.90  & 1.05 & 88.8 & 97.6 & 94.4 & 82.0 & 90.7 \\
        & $\tau=1e^{-3}$ & 6.62  & 0.90 & 89.0 & 98.4 & 94.6 & 83.8 & 91.5 \\
        & $\tau=5e^{-3}$ & 4.26  & 0.42 & 84.2 & 94.4 & 90.6 & 62.2 & 82.9 \\
        & $\tau=1e^{-2}$ & 3.36  & 0.19 & 76.2 & 87.2 & 81.6 & 42.6 & 71.9 \\

        \midrule[0.5pt]

        \multirow{6}{*}{\textit{\shortstack{Pure KL\\(Threshold)}}} 
        & $\tau=1e^{-4}$ & 10.58 & 1.40 & 91.2 & 98.2 & 93.8 & 81.8 & 91.3 \\
        & $\tau=2e^{-4}$ & 9.25  & 1.12 & 92.0 & 98.6 & 95.0 & 82.0 & \textbf{91.9} \\
        & $\tau=5e^{-4}$ & 7.66  & 0.95 & 90.8 & 99.4 & 93.2 & 82.0 & 91.4 \\
        & $\tau=1e^{-3}$ & 6.57  & 0.79 & 89.8 & 98.2 & 93.8 & 79.6 & 90.4 \\
        & $\tau=5e^{-3}$ & 4.30  & 0.40 & 86.2 & 91.4 & 87.2 & 50.8 & 78.9 \\
        & $\tau=1e^{-2}$ & 3.38  & 0.18 & 75.8 & 82.2 & 81.8 & 33.8 & 68.4 \\

        \bottomrule[1pt]
    \end{tabular}
    }
\end{table}

\begin{figure}[t]
    \centering
    \includegraphics[width=\linewidth]{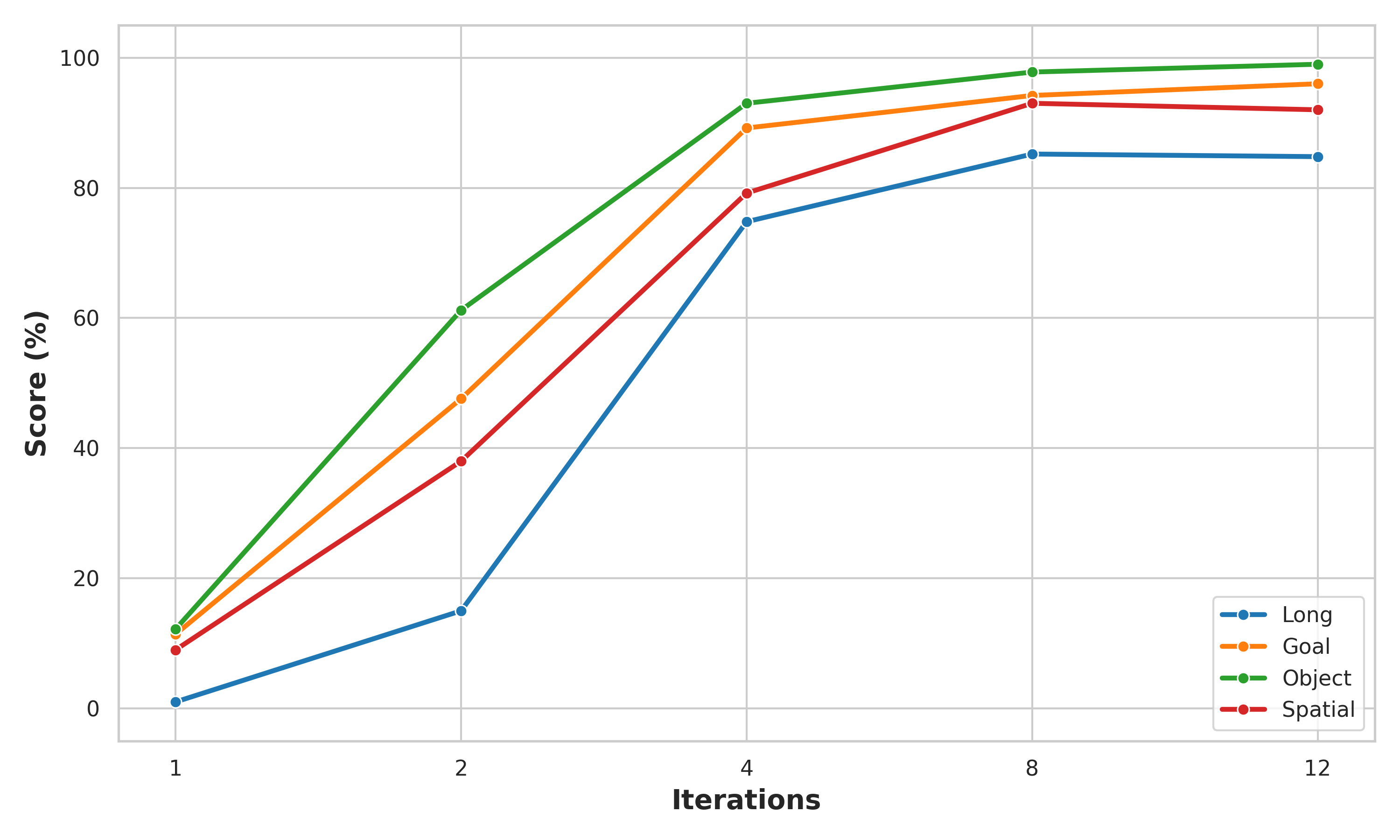}
    \caption{Performance across LIBERO benchmarks for different numbers of recurrences. All task categories show consistent improvement with increased computational depth, with models converging between 8--12 iterations on average.}
    \label{fig:main}
\end{figure}

We first establish the relationship between computational depth and task success rates by evaluating RD-VLA with a fixed number of recurrent iterations $N_{inf} \in \{1, \dots, 32\}$ on all task suites of LIBERO. As shown in Figure~\ref{fig:main}, performance exhibits a clear log-linear improvement with increased recurrence. Starting from near-random performance at $N_{inf}=1$ (8.4\% average), the model achieves substantial gains at each doubling of compute: 40.5\% at $N_{inf}=2$ (382\% increase), 84.1\% at $N_{inf}=4$ (108\% increase), and 92.6\% at $N_{inf}=8$ (10\% increase). Performance saturates between 8 and 12 iterations, with the peak of 93.1\% achieved at $N_{inf}=24$. Notably, scaling beyond $N_{inf}=12$ yields diminishing returns, suggesting that for the given task distribution, the model shows an increasing trend of its performance on different tasks by scaling up the compute budget at test-time.

\subsection{Necessity of Task-dependent Computation}
\label{subsec:task_dependent}

\begin{figure}[t]
    \centering
    \includegraphics[width=1.\linewidth]{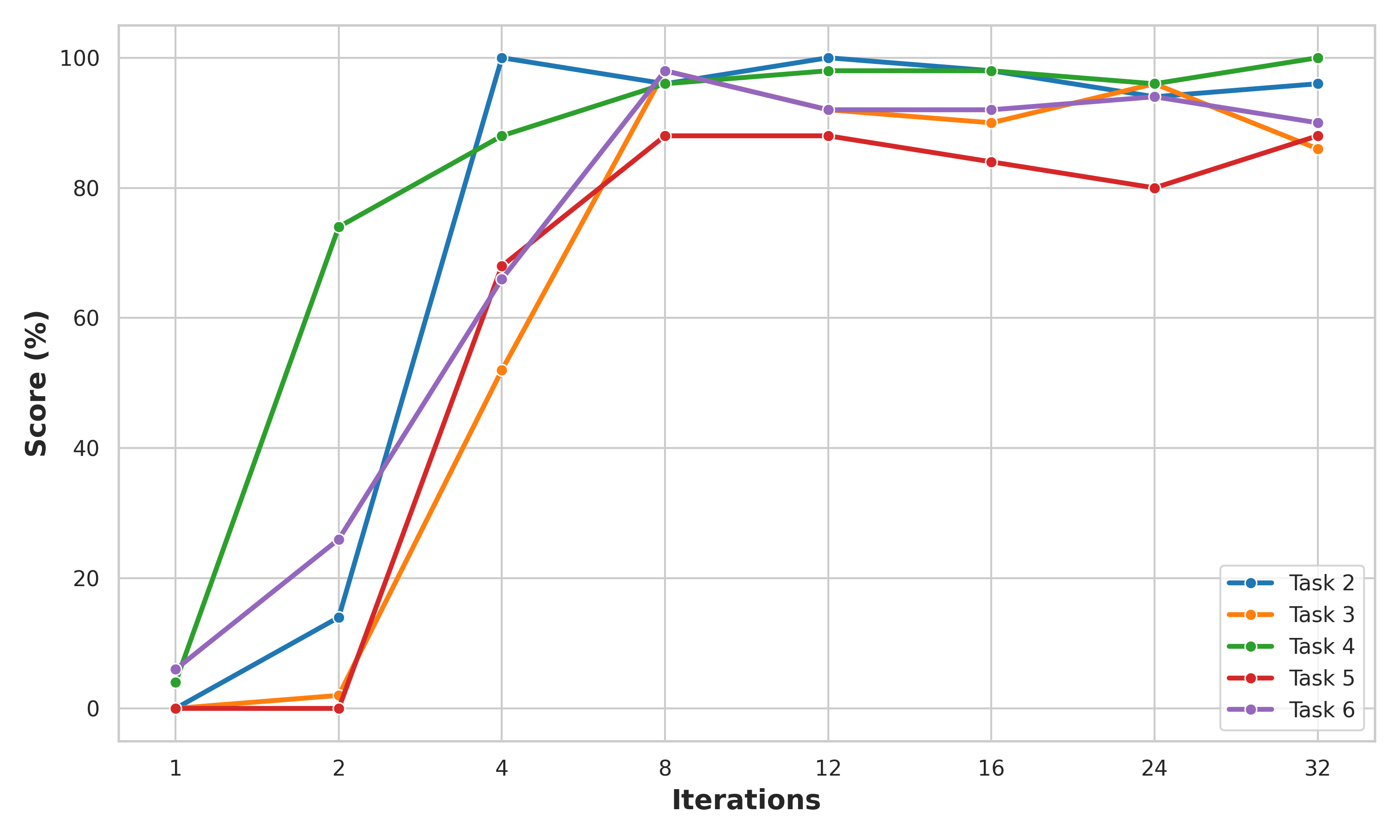}
    \caption{Performance on selected 5 Long tasks across recurrence steps. Each task exhibits distinct convergence behavior: Task 4 jumps from 6\% at iteration 1 to nearly 80\% at iteration 2, while Task 5 remains at 0\% through iteration 2 and only reaches $\sim$70\% at iteration 3. This demonstrates the task-dependent and emergent adaptive behavior of our model.}
    \label{fig:task_convergence}
\end{figure}

Beyond aggregate performance, we observe that individual tasks exhibit distinct convergence profiles, reflecting that different tasks have different requirements for computation complexity. Figure~\ref{fig:task_convergence} illustrates this task-dependent behavior on selected Long-horizon tasks from LIBERO.

Critically, the number of required iterations emerges naturally from the task context rather than being prescribed. Some tasks (e.g., Task 4) achieve near-perfect performance with just two iterations, while others (e.g., Task 5) require three or more iterations before any meaningful success. This observation reveals the phenomenon that different tasks might have different optimal iteration counts in the latent reasoning process. This further motivates our adaptive computation strategy: rather than using a fixed compute budget, the model should dynamically allocate iterations based on the difficulty of the current state.

\subsection{Adaptive Computation Strategies}
\label{subsec:adaptive_experiments}

We evaluate whether the model can dynamically calibrate its own compute budget using the KL-divergence stopping criterion described in Section~\ref{subsec:adaptive_computation}. Table~\ref{FullAblationTable} presents comprehensive ablation studies comparing fixed computational depths against three adaptive strategies: Binary Adaptation, Linear Decay, and Pure KL thresholding.

\begin{figure*}[t]
    \centering
    \includegraphics[width=0.99\textwidth]{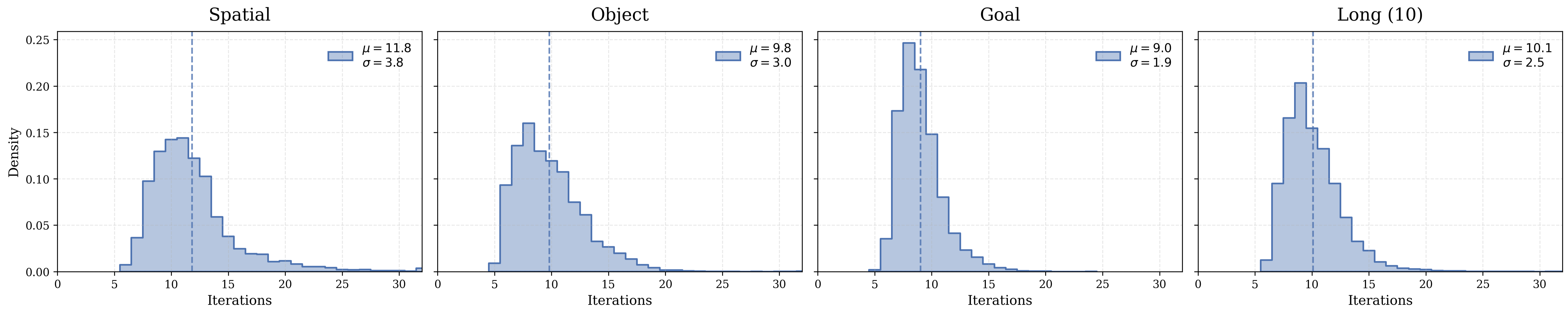} 
    \caption{Histograms of zero-shot, per-token adaptive exits based on KL divergence between consecutive steps ($\tau=10^{-4}$). The distributions show task-dependent iteration counts: Spatial tasks require more iterations ($\mu=11.8$) than Object ($\mu=9.8$) or Goal ($\mu=9.0$) tasks, reflecting their relative complexity.}
    \label{fig:KL_main}
\end{figure*}

Results demonstrate that adaptive computation maintains parity with the best fixed-depth models while reducing average inference cost. With $\tau=5 \times 10^{-4}$, the Binary Adaptation strategy achieves 92.5\% success rate (comparable to the 93.0\% peak of fixed recurrence at $N=12$) using an average of only $\bar{k}=7.93$ iterations, which is a \textbf{34\% reduction} in compute. Figure~\ref{fig:KL_main} illustrates the distribution of adaptive exits across task categories, showing that the model dynamically adjusts the number of recurrent steps based on the difficulty of the current state.

Notably, all three adaptive strategies perform comparably at matched compute budgets, suggesting that the key insight is not the specific stopping criterion but rather the principle of condition-dependent computation allocation. However, Binary Adaptation with $\tau=5 \times 10^{-4}$ achieves the best balance between efficiency and performance.

\subsection{Performance against Other Baselines}
\label{subsec:benchmark_comparison}

We compare RD-VLA against state-of-the-art VLA methods on the LIBERO and CALVIN benchmarks. Table~\ref{ComparisonLIBERO} presents results on LIBERO, where methods are grouped into three categories: end-to-end VLAs, token-level reasoning methods, and our latent reasoning approach.

RD-VLA achieves state-of-the-art performance of \textbf{93.0\%} with fixed recurrence, significantly outperforming all prior methods, including the strong Fast-ThinkAct baseline (89.7\%). Remarkably, our model achieves this with only \textbf{0.5B parameters} which is 14$\times$ smaller than 7B token-reasoning methods. The adaptive variant maintains competitive performance (92.5\%) while providing the efficiency benefits of dynamic compute allocation.

Table~\ref{ComparisonCALVIN} presents results on the CALVIN ABC$\to$D benchmark, which evaluates long-horizon task chaining. RD-VLA achieves the highest average chain length of \textbf{3.39}, outperforming OpenVLA \cite{kim2024openvla} (3.27) and all other baselines. This validates that latent refinement effectively extends the model's sequential planning capabilities.

\begin{table}[!t]
    \centering
    \small
    \setlength{\tabcolsep}{2.2mm}
    \caption{Comparison on the CALVIN ABC$\to$D benchmark. We report tasks completed in a row ($\uparrow$) and average episode length ($\uparrow$).}
    \label{ComparisonCALVIN}
    \resizebox{\linewidth}{!}{
    \begin{tabular}{l c ccccc c}
        \toprule
        \textbf{CALVIN ABC$\to$D} &
        \textbf{Params} &
        \multicolumn{5}{c}{\textbf{Tasks completed in a row $\uparrow$}} &
        \textbf{Avg. len $\uparrow$} \\
        \cmidrule(lr){3-7}
        & & 1 & 2 & 3 & 4 & 5 & \\
        \midrule

        OpenVLA\cite{kim2024openvla}        & 7   & 91.3 & 77.8 & 62.0 & 52.1 & 43.5 & 3.27 \\
        VLAS \cite{zhao2025vlasvisionlanguageactionmodelspeech}          & 7   & 87.2 & 64.2 & 40.9 & 28.1 & 19.6 & 2.40 \\
        LCB  \cite{shentu2025llmsactionslatentcodes}          & 7   & 73.6 & 50.2 & 28.5 & 16.0 &  9.9 & 1.78 \\
    
        \midrule

        DeeR  \cite{yue2025deer}         & 3   & 86.2 & 70.1 & 51.8 & 41.5 & 30.4 & 2.82 \\
        RoboFlamingo \cite{li2024visionlanguagefoundationmodelseffective}  & 3   & 82.4 & 61.9 & 46.6 & 33.1 & 23.5 & 2.48 \\
        GR-1 \cite{wu2023unleashinglargescalevideogenerative}          & 3.1 & 85.4 & 71.2 & 59.6 & 49.7 & 40.1 & 3.06 \\
        SuSIE  \cite{black2023zeroshotroboticmanipulationpretrained}        & 1.3 & 87.0 & 69.0 & 49.0 & 38.0 & 26.0 & 2.69 \\
        \midrule

        RT-1 \cite{brohan2023rt1roboticstransformerrealworld}          & --  & 53.3 & 22.2 &  9.4 &  3.8 &  1.3 & 0.90 \\
        \rowcolor[rgb]{0.90,0.90,0.90}
        \textbf{RD-VLA (Ours)}  & \textbf{0.5}  & \textbf{91.4} & \textbf{79.5} & \textbf{67.9} & \textbf{54.9} & \textbf{45.3} & \textbf{3.39} \\
        \bottomrule
    \end{tabular}}
\end{table}

\subsection{Real-world Experiments}
\label{subsec:real_world}

\begin{figure}[t]
    \centering
    \includegraphics[width=1.0\linewidth]{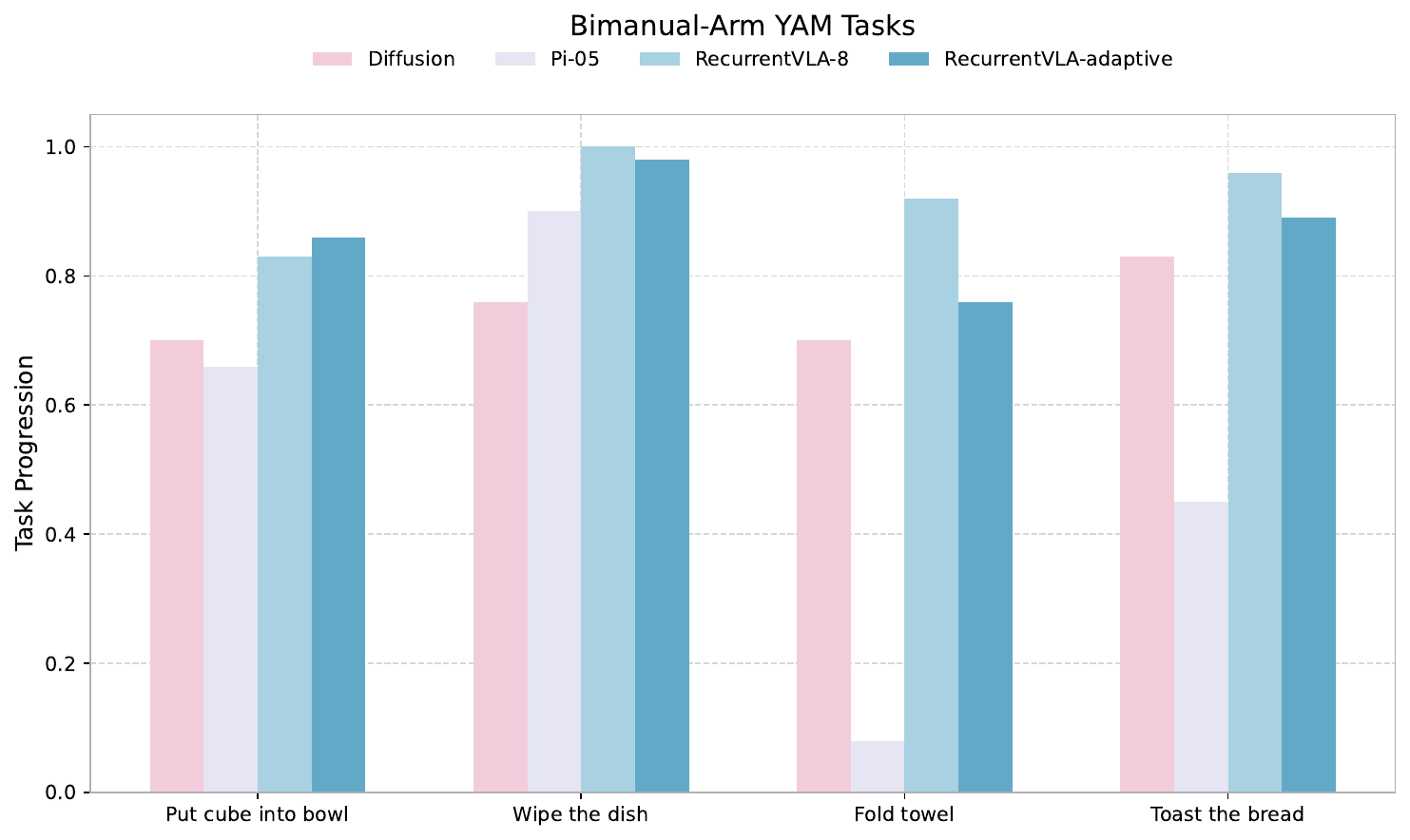}
    \vspace{-7pt}
    \begin{subfigure}{0.24\linewidth}
        \centering
        \includegraphics[width=\linewidth, keepaspectratio=True]{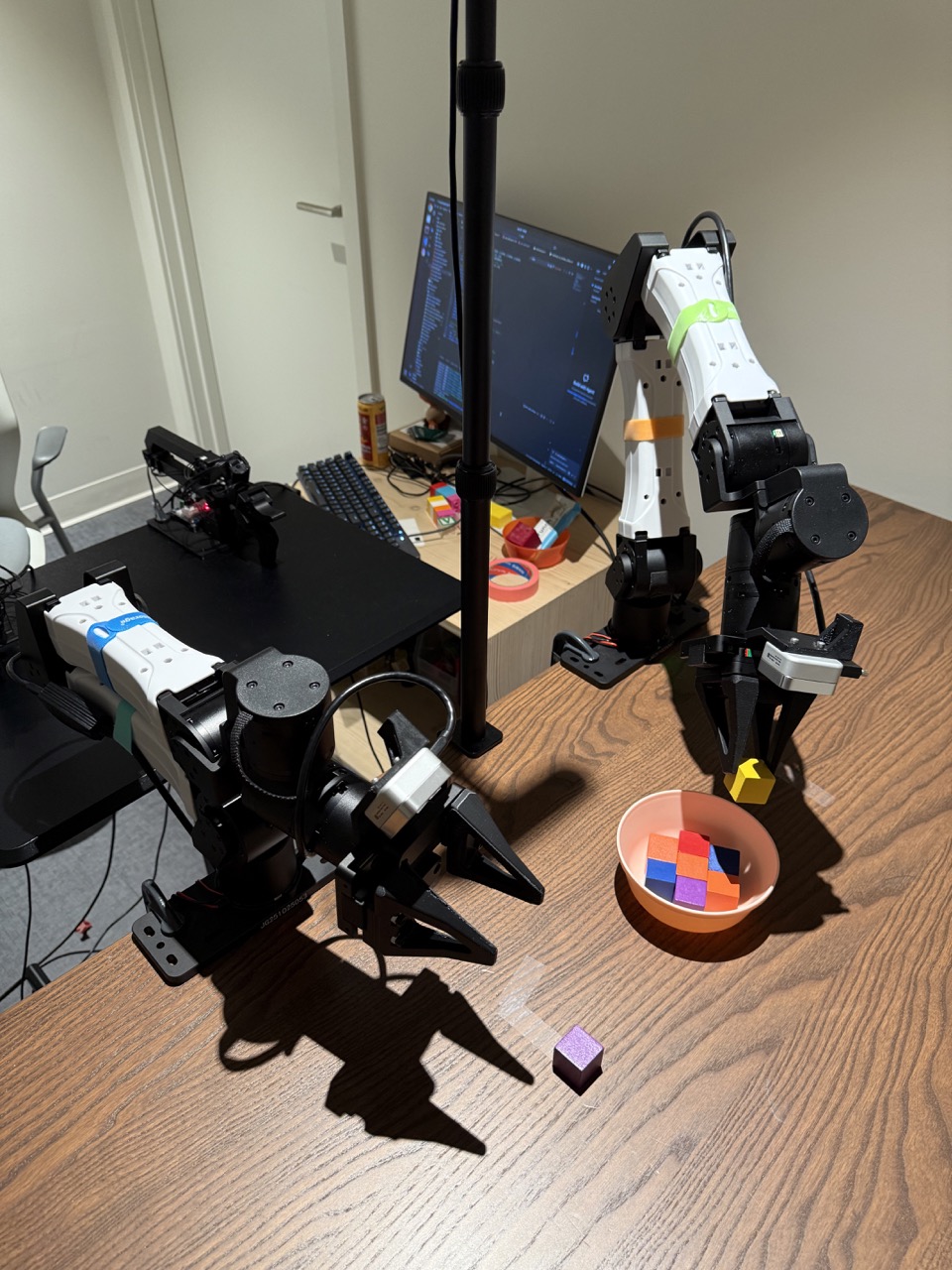}
        \caption*{\scriptsize ``Put cube into bowl"}
    \end{subfigure}
    \hfill
    \begin{subfigure}{0.24\linewidth}
        \centering
        \includegraphics[width=\linewidth, keepaspectratio=True]{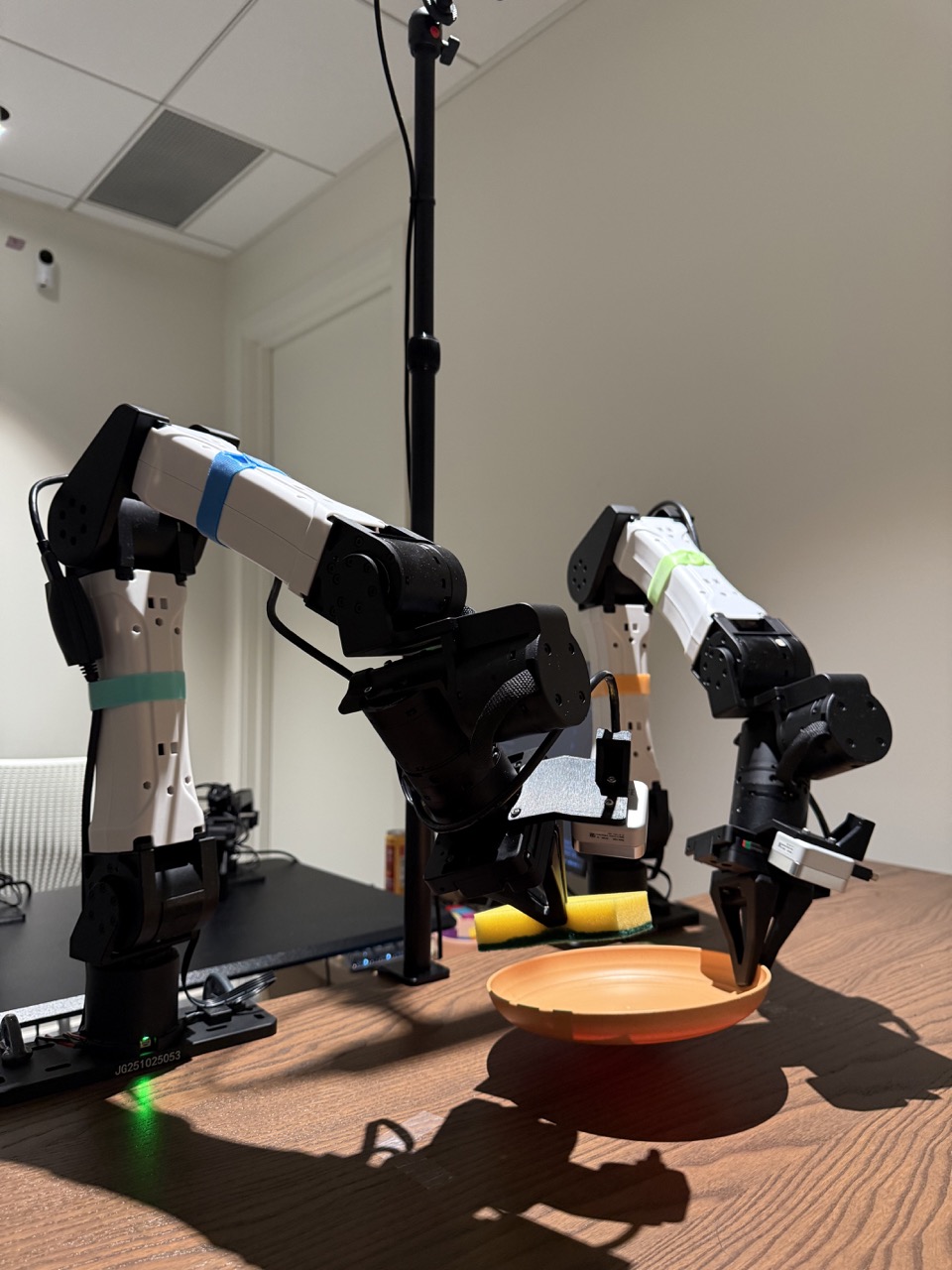}
        \caption*{\scriptsize ``Wipe the dish"}
    \end{subfigure}
    \hfill
    \begin{subfigure}{0.24\linewidth}
        \centering
    \includegraphics[width=\linewidth, keepaspectratio=True]{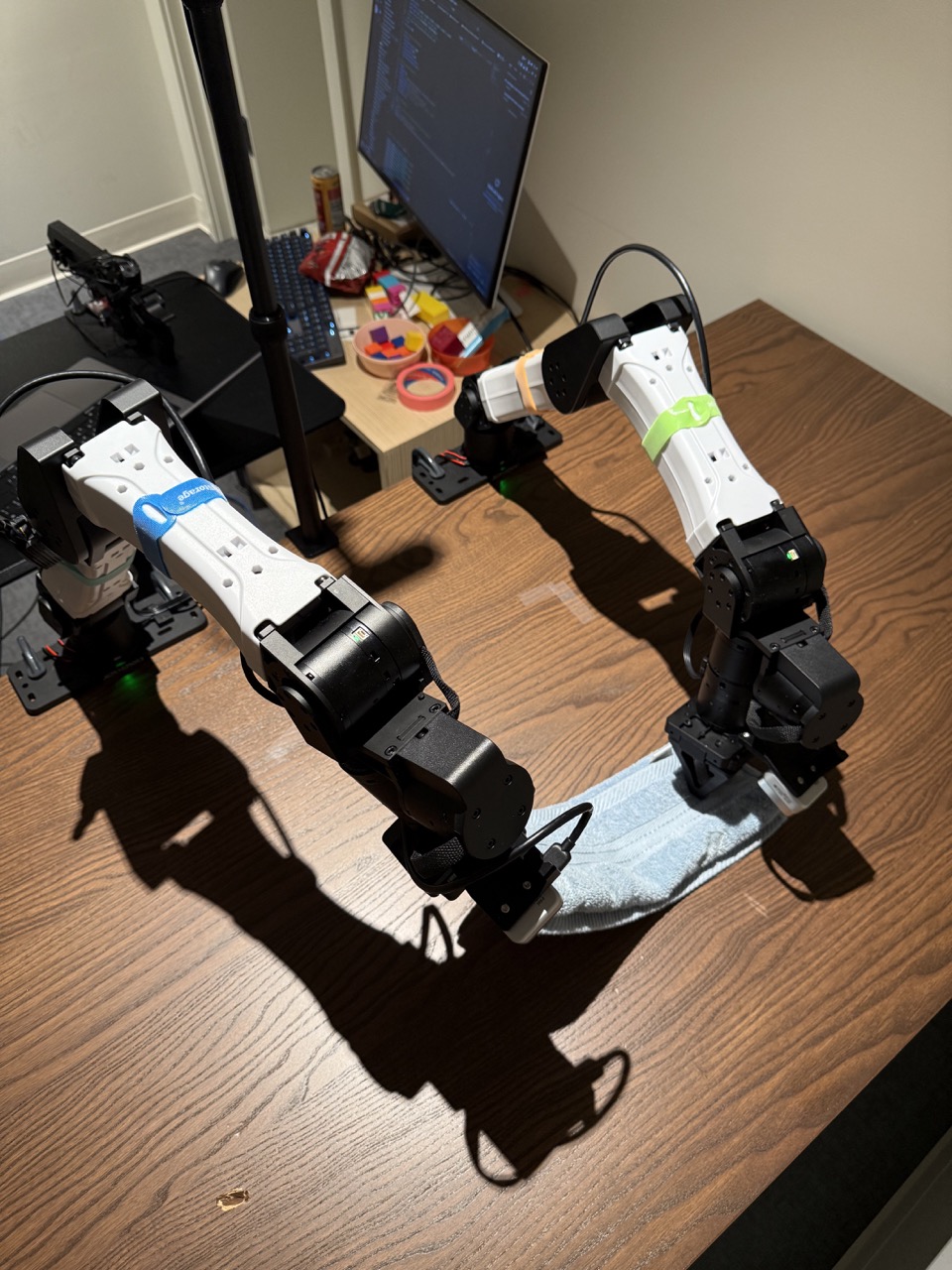}
        \caption*{\scriptsize ``Fold towel"}
    \end{subfigure}
    \hfill
    \begin{subfigure}{0.24\linewidth}
        \centering
        \includegraphics[width=\linewidth, keepaspectratio=True]{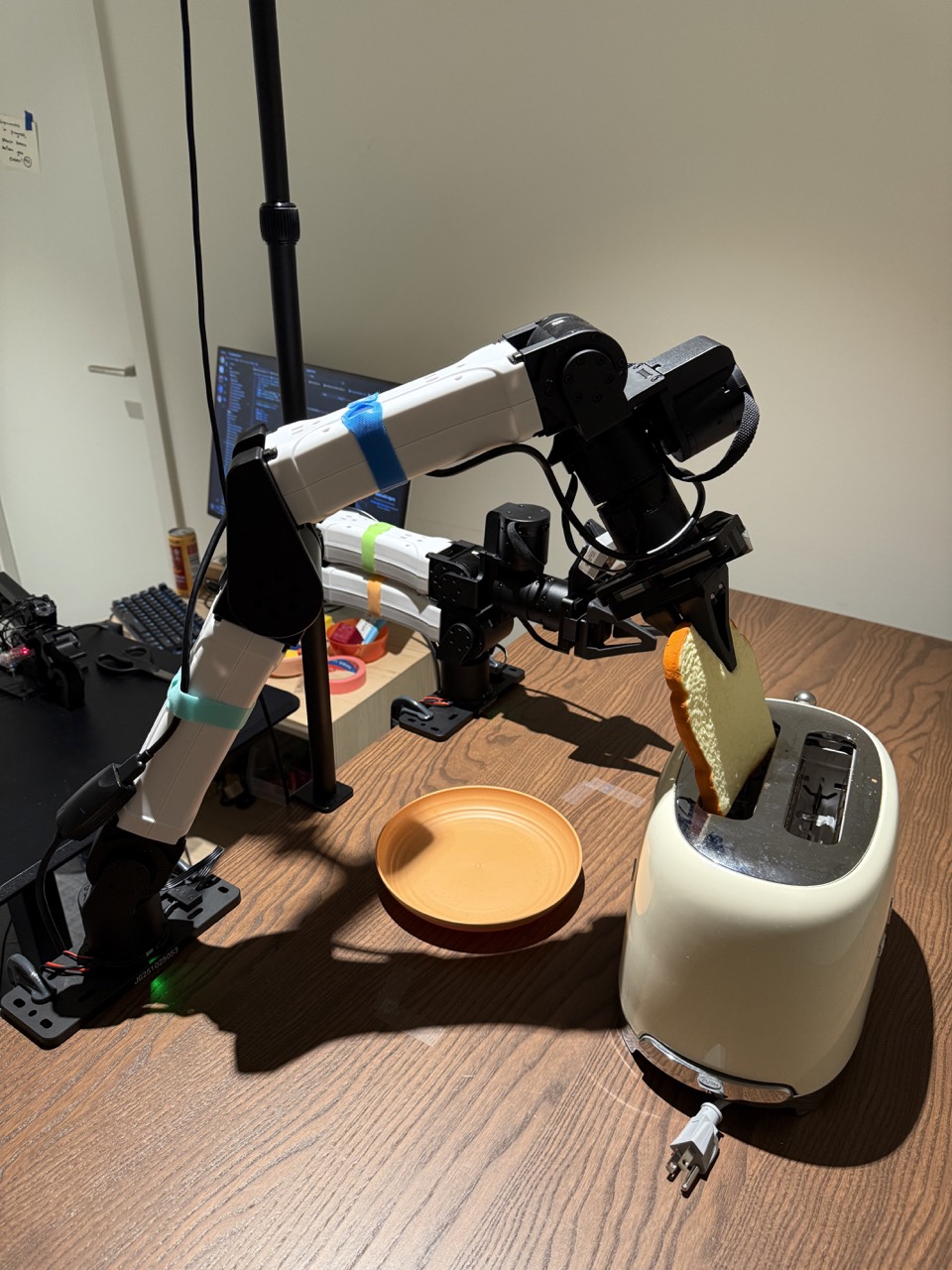}
        \caption*{\scriptsize ``Toast the bread"}
    \end{subfigure}
    \vspace{-5pt}
    \caption{Performance on real-world tasks compared to $\pi_{0.5}$ and Diffusion Policy baselines. RD-VLA variants consistently outperform baselines across all tasks, with the fixed 8-iteration model achieving near-perfect performance on dish wiping.}
    \label{fig:real}
\end{figure}

To validate the capability of our method in the real-world setting, we deployed RD-VLA on a bimanual robot arm (bimanual YAM) across four household tasks: placing a cube in a bowl, wiping a dish, folding a towel, and toasting bread. We train and compare our model in each task, both using fixed 8 iterations and Pure KL with threshold $\tau=10^{-4}$. Figure~\ref{fig:real} presents task progression scores compared to Diffusion Policy and $\pi_{0.5}$ baselines. RD-VLA with fixed 8 iterations demonstrates superior robustness across all tested scenarios, reaching nearly 100\% completion on dish wiping and significantly outperforming baselines on the remaining tasks. The adaptive variant (RD-VLA-adaptive, Pure KL with threshold $\tau=10^{-4}$) maintains competitive performance, matching or closely trailing the fixed-strategy model while achieving the highest score on cube placement. This demonstrates the viability of dynamic computation in physical settings, even when slightly trading off performance on complex manipulation tasks like towel folding.

%% file: section/conclusion.tex
\section{Discussion and Limitations}
Our primary objective was to investigate latent iterative reasoning in robotic control, rather than to hyper-optimize a specific model for state-of-the-art dominance. While RD-VLA achieves competitive or superior performance on standard benchmarks such as LIBERO \cite{liu2023liberobenchmarkingknowledgetransfer}, we emphasize that these results were obtained with minimal hyperparameter tuning and a relatively small backbone (0.5B parameters). We expect that scaling this architecture to larger backbones and training on more diverse datasets will yield significant performance gains.

A key limitation observed in our experiments is the boundary of depth generalization. While performance scales predictably with the number of recurrent steps up to some optimal number of iterations, extending recurrence beyond this number of iterations may lead to state saturation or performance degradation rather than continued refinement. Addressing this problem—perhaps through architectural innovations or specific training protocols remains an open challenge for scaling latent reasoning in robotics.

Recurrent architectures inherently expose internal state dynamics that can serve as proxies for model uncertainty. This offers a suite of test-time intervention capabilities, such as adaptive computation or uncertainty-aware execution. For instance, the system could autonomously halt execution or request operator assistance if the variance between recurrent states exceeds a safety threshold. While we demonstrated the viability of these mechanisms, the design space for such interventions is vast. We leave the specific implementation of such mechanisms to future work, focusing here on the architectural foundation that makes them possible.

While this work highlights the latency and memory limitations of token-based Chain-of-Thought (CoT) reasoning for high-frequency control, our Recurrent-Depth approach offers a complementary pathway. Future research could investigate hybrid approaches where recurrent depth is modulated per-token to enhance CoT reasoning capabilities in embodied agents.
\section{Conclusion and Future Work}
We introduced \textbf{Recurrent-Depth VLA (RD-VLA)}, a novel architecture that shifts robotic reasoning from the discrete output space to the continuous latent space. This work serves as the first implementation of \textit{latent iterative reasoning} for visuomotor policies, demonstrating that effective test-time compute scaling can be achieved without the memory and latency overhead of autoregressive Chain-of-Thought generation. Our experiments provide substantial evidence that the model successfully utilizes recurrent iterations to refine its internal state, with performance scaling log-linearly with compute depth. Crucially, this architecture unlocks new capabilities for embodied agents: the ability to think longer for harder tasks and the capacity to measure its own uncertainty through latent convergence. We aimed to open a new design space for efficient, reasoning-capable robotic policies. We believe that optimizing the regimes for adaptive compute and exploring the scaling laws of latent recurrence represent promising avenues for future research. By validating the efficacy of iterative latent refinement, RD-VLA provides a foundation for the next generation of resource-efficient and robust foundation models in robotics.

%% file: section/appendix.tex
% ============================================================
%  APPENDIX — single-column layout
% ============================================================
% REQUIREMENTS in your preamble:
%   \usepackage{float}       % provides [H] placement specifier
%   \usepackage{subcaption}  % for subfigures
%
% If your main paper uses two-column (e.g. \documentclass[twocolumn]{article}),
% the \onecolumn command below switches the appendix to single-column.
% ============================================================

\onecolumn   % <-- switches from two-column to single-column for the appendix